\documentclass[letterpaper, 10 pt, conference]{ieeeconf}  

\IEEEoverridecommandlockouts                              

\usepackage{graphicx}
\usepackage{amsmath,amssymb,amsfonts}
\usepackage{svg}
\usepackage{multirow}

\usepackage[skip=1ex, font=small]{caption}

\usepackage{color}
\usepackage[normalem]{ulem} 
\definecolor{visnecurugu}{rgb}{0.6,0.1,0.6}

\title{\LARGE \bf
Multimodal Detection and Identification of Robot Manipulation Failures
}

\author{Arda Inceoglu$^{1}$, Eren Erdal Aksoy$^{2}$, Sanem Sariel$^{1}$ 
\thanks{$^{1}$Artificial Intelligence and Robotics Laboratory, Faculty of Computer and Informatics Engineering,
      Istanbul Technical University, Maslak, Turkey 
      {\tt\small {\{inceoglua, sariel\}@itu.edu.tr}}%
      }
\thanks{$^{2}$School of Information Technology, Center for Applied Intelligent Systems Research, Halmstad  University, Halmstad, Sweden
    }
\thanks{This research is also supported by a grant from the Scientific and Technological Research Council of Turkey (TUBITAK), Grant No. 119E-436.}
}

\begin{document}
\maketitle
\thispagestyle{empty}
\pagestyle{empty}

\begin{abstract}

An autonomous service robot should be able to interact with its environment safely and robustly without requiring human assistance. Unstructured environments are challenging for robots since the exact prediction of outcomes is not always possible. Even when the robot behaviors are well-designed, the unpredictable nature of physical robot-object interaction may prevent success in object manipulation. Therefore, execution of a manipulation action may result in an undesirable outcome involving accidents or damages to the objects or environment. Situation awareness becomes important in such cases to enable the robot to (i) maintain the integrity of both itself and the environment, (ii) recover from failed tasks in the short term, and (iii) learn to avoid failures in the long term. For this purpose, robot executions should be continuously monitored, and failures should be detected and classified appropriately.
In this work, we focus on detecting and classifying both manipulation and post-manipulation phase failures using the same exteroception setup. We cover a diverse set of failure types for primary tabletop manipulation actions. In order to detect these failures, we propose FINO-Net \cite{Inceoglu2021fino}, a deep multimodal sensor fusion based classifier network. Proposed network accurately detects and classifies failures from raw sensory data without any prior knowledge. In this work, we use our extended FAILURE dataset \cite{Inceoglu2021fino} with 99 new multimodal manipulation recordings and annotate them with their corresponding failure types. FINO-Net achieves 0.87 failure detection and 0.80 failure classification F1 scores. Experimental results show that proposed architecture is also appropriate for real-time use.

\end{abstract}

\section{INTRODUCTION}

\begin{figure}[t!]
\centering
    \includegraphics[width=\linewidth]{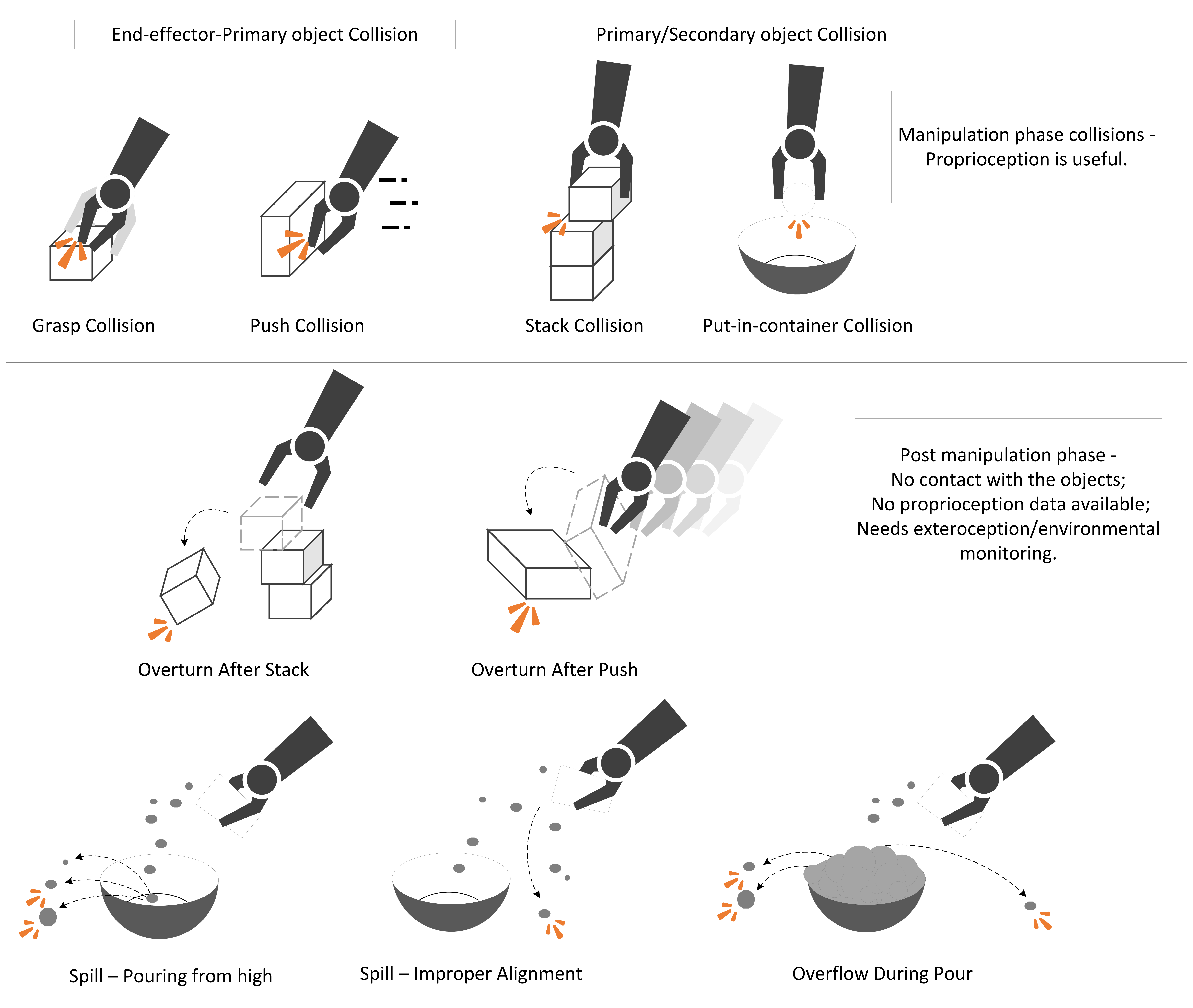}
    \caption{Manipulation failures can be categorized along two dimensions: (i) one for the collision time: either at the manipulation phase or at the post-manipulation phase; and (ii) for the object types that are in collision: either end-effector-primary object collisions or primary-secondary object collisions. Failure types on a set of primary manipulation actions (e.g., \textit{pick}, \textit{push}, \textit{stack},\textit{ put-in-container}, \textit{pour}) that we focus on in this study are categorized along these dimensions.} 
    \label{fig:fig11}
\end{figure}

Currently, robots are not robust enough to safely handle all house chores on their own, especially when executing manipulation tasks in unstructured environments where they interact with everyday objects and humans \cite{Ersen2017}. It is hard to exactly predict the actual outcomes of actions in these settings where unintended or harmful consequences may arise. This is due to the fact that, the estimations on manipulation/interaction parameters leading to success may be wrong due to incomplete or incorrect internal representation of the world; moreover, environmental factors (e.g., external events) may also prevent proper action performance, leading to undesirable outcomes.

In such circumstances, ensuring the reliability and safety of the operations of the robot is crucial \cite{veruggio2016roboethics}.

In industrial settings, safety measures are taken by external safeguarding devices, and certain regulations and standards are applied by operators \cite{iso_10218:2011_part1, iso_10218:2011_part2}. However, for autonomous operations in human workspaces, the robot itself is responsible for the safety of its actions. Situation awareness capabilities are required to make it able to effectively operate in unstructured environments without assistance \cite{morin2011self, nitsch2013situation,dos2020situational}. 
In this study, we focus on these on-board situation awareness capabilities for service robots.

If the robot can detect and classify a failure effectively, it can react appropriately when confronted with it \cite{ak2019stop}. 
This can be achieved through real-time monitoring of the robot of its workspace during manipulation. In this article, we present FINO-Net to sense presence of failures by multimodal - exteroceptive sensors without integrating any domain/task specific knowledge even though the robot knows which task it is executing. Previously failure detection \cite{Inceoglu2018icra, diehl2022did} and identification problems \cite{altan2021went} were addressed at symbolic level \cite{inceoglu2018violet}. In this study, we present FINO-Net as an data driven framework for monitoring. Egocentric vision, audition are separately used in scene analysis \cite{min2019collision}, \cite{xiaoran2021acoustic}, \cite{valle2022real}, \cite{Saltali2016}. In this work, we integrate egocentric vision with audition to complement each other.

Whilst, there is a large body of research literature addressing collision detection both in physical Human-Robot Interaction (pHRI) and manipulation tasks; most of this focuses on collision detection using only proprioceptive sensors \cite{haddadin2017robot}. Microphones, torque/force sensors, and inertial measurement units (IMU) are used to detect physical collisions with objects in manipulation settings \cite{valle2022real}. Joint/end-effector and object/human collisions are of particular interest in these works. On the contrary, we primarily focus on the post-manipulation failure phase \cite{proper2023aim} to classify a manipulation failure affecting either the primary objects being manipulated or the secondary objects that are not in direct contact with the end-effector (Figure \ref{fig:fig11}). Effective post-manipulation failure detection requires environmental monitoring and the use of exteroceptive sensors since a failure may happen in the post-impact phase (i.e., end-effector and object contact ends). We focus on how this kind of detection can be performed by using a minimal set of onboard sensors placed on top of the robot.

We focus on failures that may occur during real-time execution of primitive object manipulation actions where the robot end-effector is in close contact (i.e., without tools) with manipulated objects on the tabletop. Failure types that we consider are: collisions with objects (with manipulated or secondary objects), missing objects being manipulated, overflowing or spilling the contents of a container, and overturning of objects. The causes of these failures are as follows:

\begin{itemize}
    \item Improper alignment of the robot arm with respect to the manipulated object due to misspecified or misidentified parameters (e.g., grasping or pushing from a misaligned orientation, pouring from high, etc.) based on the internal representations of the world,
    \item Incorrect estimations or perceptual errors on objects (e.g., underestimating the contents of a container),
    \item Wrong assumptions on the skills of the self. 
\end{itemize}

Perception has a crucial role in detecting and identifying failures. Furthermore, multisensory integration is more desirable to take advantage of each sensor's perceptual contribution \cite{Inceoglu2021fino}. However, the design and development of high-quality perception systems require large amounts of high-quality data. Recently, there have been great efforts to collect large-scale robot manipulation data \cite{Mandlekar2018roboturk, Mandlekar2019roboturk, Dasari2019robonet, Levine2018learning, Finn2016unsupervised}. These efforts, however, center around manipulation skill learning tasks and ignore failures emerging during manipulation executions. To the best of our knowledge, there are no publicly available multimodal datasets on robot manipulation failures.

The FAILURE \cite{Inceoglu2021fino} dataset rather focuses on the failed attempts of various robot manipulation actions (e.g., push, pour, pick\&place, etc.) while collecting multimodal sensor readings such as RGB images, depth data, and audio waves.
It covers a variety of manipulation failures that are emerged from different affected objects or at different manipulation phases. Figure \ref{fig:fig11} depicts a categorization of failure types. A failure can occur during the manipulation phase or in the post-manipulation phase. On the other hand, the end-effector may involve in collision with the primary object, or the primary and secondary objects may collide with each other. In this work, we address all these interactions during manipulation and post-manipulation phase.

To address the above mentioned challenges, we present FINO-Net \cite{Inceoglu2021fino}, a deep sensor fusion-based multimodal classifier network to detect manipulation failures. FINO-Net detects manipulation failures effectively by combining visual (RGB and depth) and auditory modalities. In addition, our network uses early fusion to combine RGB and depth frames, while late fusion is used to combine vision and audio data. To represent spatio-temporal features in sensory observations, modalities are processed separately with a series of convolutional and convolutional-LSTM layers, and the latent space representations are combined to detect potential failures. In this work, we extend FINO-Net to further classify manipulation failure types. Furthermore, we deeply analyze its real-time on demand failure detection and classification capabilities on the extended dataset.

Contributions of this paper are as follows: (i) We present the multimodal sensor fusion based FINO-Net architecture. (ii) FINO-Net architecture is analyzed for real-time on demand failure detection and classification. (iii) FAILURE dataset is extended with 99 new manipulation action recordings. (iv) FAILURE dataset is annotated with failure types. 

Only a single manipulation action (e.g., grasping) is focused in existing execution monitoring and failure detection studies. Their main focus is on robot-primary object interactions. In comparison, we cover a set of primary  manipulation actions. Furthermore, we also address failures involving both primary (i.e., manipulated main object) and secondary object (i.e., other objects with which the robot is not in direct contact) in the workspace.

\section{Related Work}

In the literature, \textit{failure (a.k.a. fault or anomaly) detection} and \textit{execution monitoring} keywords are used interchangeably. The work in \cite{Chalapathy2019} summarizes deep learning-based anomaly detection for various application domains. From the robotics perspective, there are model-based and model-free approaches proposed for execution monitoring \cite{Gertler1998}. The former approach compares the already known models with observations, whereas the latter uses sensory observation to make predictions \cite{Fritz2005, Pettersson2005} \cite{Pettersson2007}.

Among recent works, \cite{Mauro2019} extends planning with a vision-based execution monitoring system, \cite{Sathish2019} analyzes different preprocessing techniques for introspective data to detect gearbox failures. Non-parametric Bayesian models \cite{Zhou2020} and Non-parametric Hidden Markov Models (HMMs) \cite{Wu2019} are investigated to detect and classify anomalies.

A multimodal execution monitoring system \cite{Park2018, Park2019} for assistive feeding task has been proposed. The authors adopt LSTM-based variational autoencoders to process multimodal input from a sensor set including a camera, a microphone, a joint encoder, and a force sensor. In another work \cite{thoduka2021}, multimodal cues are used to detect book manipulation failures on shelves. \cite{gohil2022} fuses visuo-tactile cues for grasp failure detection.

Furthermore, \cite{altan2021went} extracts predicates from multimodal inputs, which are then combined for anomaly-cause identification. In a later work \cite{altan2022}, end-to-end approaches are also investigated.

This work differs from the existing studies in that it addresses the problem of detecting and identifying manipulation failures, which are mainly caused by uncertainties in perception and execution. Unlike the works in \cite{Park2018,Park2019}, where the focus is rather on the failures emerging during human-robot interaction, this work investigates the robot-object interaction failures observed over the course of an object manipulation.

Instead of hand-crafted features such as gripper status, audio events, and object displacements used in \cite{Inceoglu2018icra,Inceoglu2018iros} or sound energy, spoon position, and mouth position used in \cite{Park2018,Park2019}, the proposed perception framework learns feature representations directly from the raw multimodal sensory data in an end-to-end fashion.

\begin{table}[t!]
\caption{Distribution of the manipulation data \label{tbl:dataset}}
\resizebox{\columnwidth}{!}{ 
\begin{tabular}{|r|ccc|ccc|c|}
\hline
\multicolumn{1}{|l|}{} &
  \multicolumn{3}{c|}{\textbf{\#Successes}} &
  \multicolumn{3}{c|}{\textbf{\#Failures}} &
  \textit{Total} \\ \hline
\textbf{Manipulation} &
  \multicolumn{1}{c|}{\textbf{Existing}} &
  \multicolumn{1}{c|}{\textbf{New}} &
  \textbf{Total Success} &
  \multicolumn{1}{c|}{\textbf{Existing}} &
  \multicolumn{1}{c|}{\textbf{New}} &
  \textbf{Total Failure} &
   \\ \hline
\textbf{Push}               & \multicolumn{1}{c|}{12} & \multicolumn{1}{c|}{18} & 30 & \multicolumn{1}{c|}{19} & \multicolumn{1}{c|}{12} & 31 & 61 \\ \hline
\textbf{Pick\&place }       & \multicolumn{1}{c|}{13} & \multicolumn{1}{c|}{2}  & 15 & \multicolumn{1}{c|}{30} & \multicolumn{1}{c|}{2}  & 32 & 47 \\ \hline
\textbf{Pour}               & \multicolumn{1}{c|}{25} & \multicolumn{1}{c|}{24} & 49 & \multicolumn{1}{c|}{42} & \multicolumn{1}{c|}{4}  & 46 & 95 \\ \hline
\textbf{Put-in-container} & \multicolumn{1}{c|}{23} & \multicolumn{1}{c|}{21} & 44 & \multicolumn{1}{c|}{31} & \multicolumn{1}{c|}{9}  & 40 & 84 \\ \hline
\textbf{Stack-a-tower}         & \multicolumn{1}{c|}{9}  & \multicolumn{1}{c|}{4}  & 13 & \multicolumn{1}{c|}{21} & \multicolumn{1}{c|}{3}  & 24 & 37 \\ \hline \hline
\multicolumn{1}{|c|}{\textit{Total}} &
  \multicolumn{1}{c|}{82} &
  \multicolumn{1}{c|}{69} &
  151 &
  \multicolumn{1}{c|}{143} &
  \multicolumn{1}{c|}{30} &
  173 &
  324 \\ \hline
\end{tabular}
}
\end{table}

\begin{table}[t!]
\caption{Distribution of Failure Types by Manipulation Action \label{tbl:dataiden}}
\resizebox{\columnwidth}{!}{ 
\begin{tabular}{|c|c|c|c|c|c||c|}
\hline
\textit{Type \textbackslash Action} & \textbf{put-in-container} & \textbf{pour}        & \textbf{push}        & \textbf{place}       & \textbf{stack-a-tower} & \textit{Total} \\ \hline
\textbf{Success}                                        & 44                        & 50                   & 30                   & 15                   & 13                     & 152                                \\ \hline
\textbf{Collision}                                      & 33                        &                      & 2                    & 23                   & 19                     & 77                                 \\ \hline
\textbf{Miss}                                           & 7                         &                      & 6                    & 7                    & 2                      & 22                                 \\ \hline
\textbf{Overflow}                                       &                           & 25                   &                      &                      &                        & 25                                 \\ \hline
\textbf{Spill}                                          &                           & 20                   &                      &                      &                        & 20                                 \\ \hline
\textbf{Overturn}                                       &                           &                      & 23                   & 2                    & 3                      & 28                                 \\ \hline \hline
\textit{Total}                                          & 84                        & 95                   & 61                   & 47                   & 37                     &   \\ \hline            
\end{tabular}
}
\end{table}

\section{Real-time Detection and Classification of Manipulation Failures}

In the following subsections, the problem description, data, and network details are presented.

\subsection{Problem Description}

Manipulation failures are inevitable in unstructured environments due to limitations in perception, estimation, and the physical capabilities of the robot. For safety reasons, robots need to be situationally aware of their actions' outcomes. Situation awareness can be formally defined as the spatio-temporal perception of the elements in the environment, their interpretation, and the projection of their state in the near future \cite{Endsley1995situation}. 

In order to build situationally aware robots, their execution should be continuously monitored, as an unexpected event may occur in any phase of the execution. An execution monitoring system should incorporate both spatial and temporal information obtained from the sensors to interpret changes in the scene during manipulation. Furthermore, monitoring the execution with multiple sensors is helpful as each sensing modality provides complementary information \cite{Inceoglu2018icra, Inceoglu2018iros, Inceoglu2021fino}.

In this work, we focus on real-time and on-board (i.e., egocentric perception with sensors mounted on the robot) detection and classification of tabletop manipulation failures caused by the robot itself. We define manipulation failure detection as the process of detecting unexpected outcomes during robot execution and failure classification as explaining the types of failures.

For addressing the aforementioned problems, failure detection and failure classification problems are modeled as classification tasks, where the inputs are multi-sensory observation sequences. and the targets are $y \in \{success, fail\}$ for detection, and $y \in \{success, collision, miss, overflow, spill, overturn \}$ for classification.

 Let $M= \bigcup_{m} (m \in \{RGB, Depth, Audio, etc. \})$ be the set of sensing modalities. 
 $x_i^m$ is a complete 
 observation sequence obtained from modality $m$ (e.g., all RGB images of a pouring action), and $i$ is the 
 manipulation recording index of the 
 multimodal observation sequence. $x_{i}^{m_{t_m}}$ represents an observation (e.g., a single RGB frame) at time step 
 $t_m$ where $t_m$ is the sampled time index for modality $m$. We construct a dataset $D$ ($|D| = N$) containing multimodal observation sequences:

\begin{equation}
D = \{ \{(x_{i}^{m_{t_1}},...,x_{i}^{m_{t_{m}}})\}_{m=1}^M, y_i\}_{i=1}^{N}
\end{equation}

The goal is to learn a function $\Phi(\cdot)$ that classifies multimodal sensory data to a label $y$: 
\begin{equation}
y = \Phi(\phi_1(x^{1_{1}},...,x^{1_{t_1}}), ... , \phi_m(x^{m_1},...,x^{m_{t_m}}))
\end{equation}

\begin{figure}[t!]
\centering
    \includegraphics[width=0.6\linewidth]{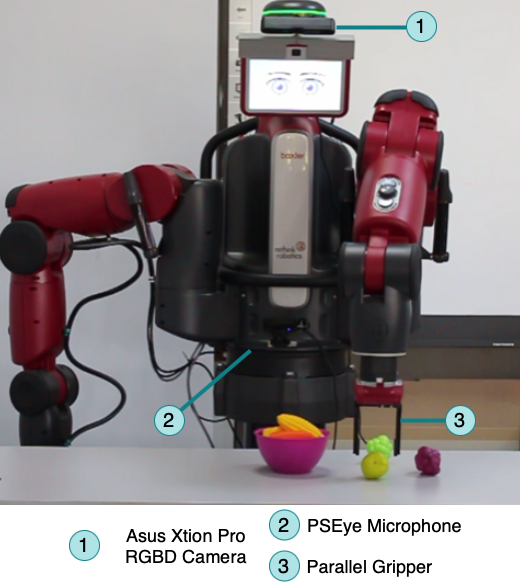}
    \caption{Experimental Environment}
    \label{fig:baxter_setup}
\end{figure}

\begin{figure*}[t!]
\centering
    \includegraphics[width=\linewidth]{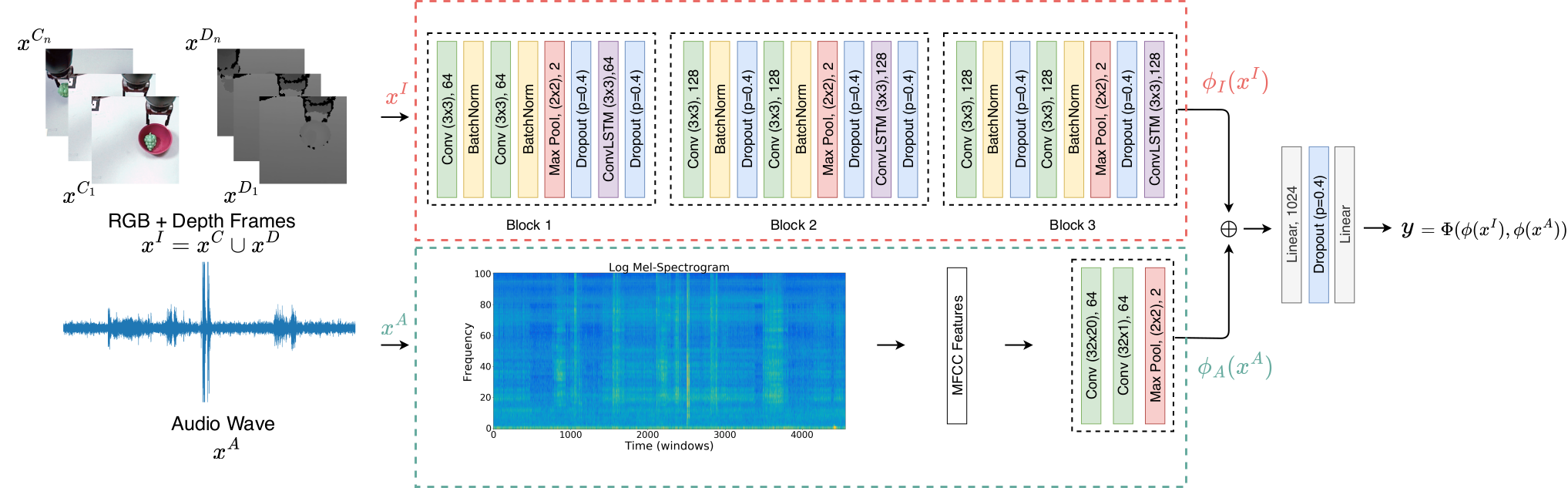}
    \caption{FINO-Net Architecture}
    \label{fig:finonet}
\end{figure*}

\subsection{Data}

In our previous work, we introduced FAILURE dataset \cite{Inceoglu2021fino}. To the best of our knowledge, there currently exists no public datasets as collections of multimodal robot execution traces for both success and fail cases covering diverse set of manipulation actions.

FAILURE dataset was constructed by using a Baxter humanoid robot equipped with the following equipments: a parallel gripper, an Asus Xtion Pro RGB-D camera mounted on the head, and a PSEye microphone mounted on the lower torso (See Figure \ref{fig:baxter_setup}). During the data collection, the robot is tasked to execute a manipulation action, and all synchronized sensor readings (i.e., RGB/RGB-D image streams, audio waves) are then simultaneously recorded.

In this work, we extend the dataset with 99 new manipulation execution recordings. Table \ref{tbl:dataset} presents the distribution of the existing and newly collected data.

Furthermore, we have annotated the recordings belonging to failure cases in the FAILURE dataset. The resulting execution status types are as follows:

\begin{itemize}
    \item \textbf{Success:} Execution is completed as expected. Minor deviations from the expected target location or orientation are tolerated.

    \item \textbf{Collision:} A collision failure occurs during all action executions except pour. In this failure type, both the manipulated object and secondary objects are affected.

    \item \textbf{Miss:} A miss failure occurs during all action executions except pour. A miss failure occurs during \textit{push} or \textit{pick-and-place} action execution. The robot either completely fails to interact with the object or fails to grasp it due to alignment errors.

    This failure type occurs when the location or size of the target object is miscalculated. 
    
    \item \textbf{Overflow:} An overflow failure occurs during the execution of a pouring action. The poured content completely fills the target container and overflows onto the table.

    \item \textbf{Spill:} A spill failure occurs during pouring action. The poured content is spilled onto the table due to alignment errors, the pouring velocity, the size mismatch between source and target containers, etc. In comparison to overflow failure, in this failure type, the target container is not full at the end of the execution.

    \item \textbf{Overturn:} An overturn failure occurs during either pushing, placing, or stacking-a-tower action. In this failure type, only the manipulated object is affected and overturned during the manipulation phase.
\end{itemize}

During failure annotation, we grouped the interactions between the robot, the target object, and the other objects in the workspace. Table \ref{tbl:dataiden} presents the distribution of failure types over manipulation actions. Note that the presented failure classes are challenging as (i) a single failure type can occur during the execution of different actions. For instance, an overturn failure can be observed during either a put-in-container, place, or stack-a-tower action; (ii) a manipulation action can result in different types of failures. For instance, during the execution of the pour action, overflow or spill failures can be observed.

\subsection{FINO-Net}

We present FINO-Net, a deep sensor fusion-based multimodal classifier network to detect manipulation failures using onboard sensory data in real time. An earlier version has appeared in \cite{Inceoglu2021fino}. This work extends FINO-Net to further classify manipulation failure types. Furthermore, we present detailed analysis on failure detection with the extended dataset.

The inputs of the network are composed of RGB ($x^C$) and depth ($x^D$) frames captured from the  head camera and audio waves ($x^A$) recorded over the course of a robot manipulation action. FINO-Net adopts early fusion ($x^I= x^C \cup x^D$) to combine RGB and depth frames while applies  late fusion ($\Phi$) to combine visual ($\phi_I$) and auditory ($\phi_A$) features. The architecture processes visual and auditory inputs individually with a series of convolutional and convolutional-LSTM (convLSTM) layers. Finally, in the fusion step, the latent space representations are concatenated into a feature vector, and fed to the fully connected layers. The overall FINO-Net architecture is depicted in Fig.~\ref{fig:finonet}. In the following subsections, we elaborate more on the network architecture.

\subsubsection{Vision ($\phi_I$)}
In the preprocessing step, 8 RGB and depth image pairs are sampled, representing the complete execution sequence. Prior to the sampling step, self-occluded frames are eliminated with a depth-based thresholding approach. The remaining samples are roughly segmented into \textit{approach}, \textit{manipulate} and \textit{retreat} phases. Then, from each \textit{approach} and \textit{retreat} phase, 4 frames are sampled.

In order to process spatio-temporal features in RGB and depth frames, convLSTM cells are employed. A typical LSTM cell is implemented using fully connected layers, while a convLSTM replaces these with convolution operators.

The visual branch consists of three main blocks (see the top branch in Fig.~\ref{fig:finonet}). Each block is composed of two convolutional layers and a convLSTM layer.
RGB and depth frames are early fused by stacking on top of each other before feeding into the first visual block. Inside each block, the filter numbers remain the same for all convolutional and convLSTM layers. Each convolutional layer has 3x3 filters. Before applying convLSTM layers, max pooling is applied to cut the number of features in half. Each block also has batch normalization and dropout layers.

\subsubsection{Audition ($\phi_A$)}
In our earlier works \cite{Inceoglu2018icra, Inceoglu2018iros} we have shown that the use of audition data helps in detecting drop, collission types of failures for which RGB and depth data do not provide sufficient clues due to occlusions, etc. Therefore, audition data was used as a complementary modality to the others. Our earlier work on audition data processing involved Mel Frequency Cepstral Coefficients (MFCCs) as a representation for auditory monitoring, where Support Vector Machines were employed to classify audio features into symbolic states: \textit{drop}, \textit{collision}, \textit{idle} and \textit{ego-noise} \cite{Inceoglu2018icra, Inceoglu2018iros}. In comparison, FINO-Net directly classifies MFCC features into execution states: \textit{success} or \textit{failure}. 

FINO-Net adopts a convolutional network composed of two convolutional layers followed by a max pooling layer (see the bottom branch in Fig.~\ref{fig:finonet}). There are 64 filters in each layer, with a filter size of 32. As input, we use single-channel audio recordings with a 16 kHz sampling rate. The raw audio signal is divided into 32 millisecond windows. For each window, Short Time Fourier transform is applied to convert the signal into the frequency domain. A Mel filterbank is applied, and 20 MFCCs are obtained. The number of audio windows is fixed by either applying padding or clipping.

\subsubsection{Fusion ($\Phi$)}
FINO-Net adopts a late fusion approach to combine visual and auditory modalities. We introduce the following model:
\begin{equation}
y = \Phi(\phi_I(x^{I_1},...,x^{I_{t_I}}) \oplus  \phi_A(x^{A_1},...,x^{A_{t_A}}))~,
\end{equation}

where $\phi_m$ is a unimodal convolutional neural network which acts as feature extractor, $\oplus$ is the concatenation operator, and $\Phi$ is the late fusion based classifier network. In the fusion step, vision and audition features, obtained from final output of each modality, are concatenated into a single feature vector. The fusion layer is composed of two fully connected layers as shown in Fig.~\ref{fig:finonet}.  

\begin{table}[t!]
\caption{Quantitative Results}
\label{tbl:quant}
\resizebox{\columnwidth}{!}{%
\begin{tabular}{|l|ccc|ccc|}
\hline
                          & \multicolumn{3}{c|}{\textbf{Detection}}    & \multicolumn{3}{c|}{\textbf{Classification}} \\ \hline
                          & \textbf{Pr} & \textbf{Re} & \textbf{F1}    & \textbf{Pr}  & \textbf{Re}  & \textbf{F1}    \\ \hline
\textbf{FINO-Net-RGB}     & 0.7667       & 0.7567       & 0.7617          & 0.6924        & 0.6216        & 0.6551          \\ \hline
\textbf{FINO-Net-D}       & 0.6898       & 0.6756       & 0.6826          & 0.7302        & 0.6486        & 0.6869          \\ \hline
\textbf{FINO-Net-RGB-D}   & 0.7817       & 0.7567       & 0.7690          & 0.6846        & 0.6756        & 0.6801          \\ \hline
\textbf{FINO-Net-A}       & 0.6801       & 0.6486       & 0.6640          & 0.7475        & 0.6216        & 0.6788          \\ \hline
\textbf{FINO-Net-RGB-D-A} & 0.8665       & 0.8648       & \textbf{0.8656} & 0.8085        & 0.7837        & \textbf{0.7959} \\ \hline
\end{tabular}%
}
\end{table}

\subsubsection{Regularization} 
Batch normalization is applied after each convolutional layer for regularization. To boost the roles of basic features (e.g., edges and curves), a central dropout approach is adopted with a probability rate of $0.4$. After each convolutional, convLSTM, and fully connected layers, a dropout layer is inserted except for the first convolutional layer in Block 1 and the last convLSTM layer in Block 3.

\section{Experiments}

In this section, we quantitatively assess the prediction performance of the FINO-Net architecture on both failure detection and classification tasks using our extended FAILURE dataset. Additionally, we analyze FINO-Net for real-time monitoring. For this purpose, we first look at the effect of sampled frames from an execution, and then, we investigate prediction accuracy with different segments of the execution.

The visual branch of the FINO-Net architecture contains three convolutional-LSTM blocks. The number of blocks is determined empirically. Reducing the number of these blocks results in lower representation capacity. Increasing the number of blocks improves accuracy slightly, but when compared to the number of parameters, three is the ideal number of blocks.

\begin{figure}[t!]
\centering
    \includegraphics[width=\linewidth]{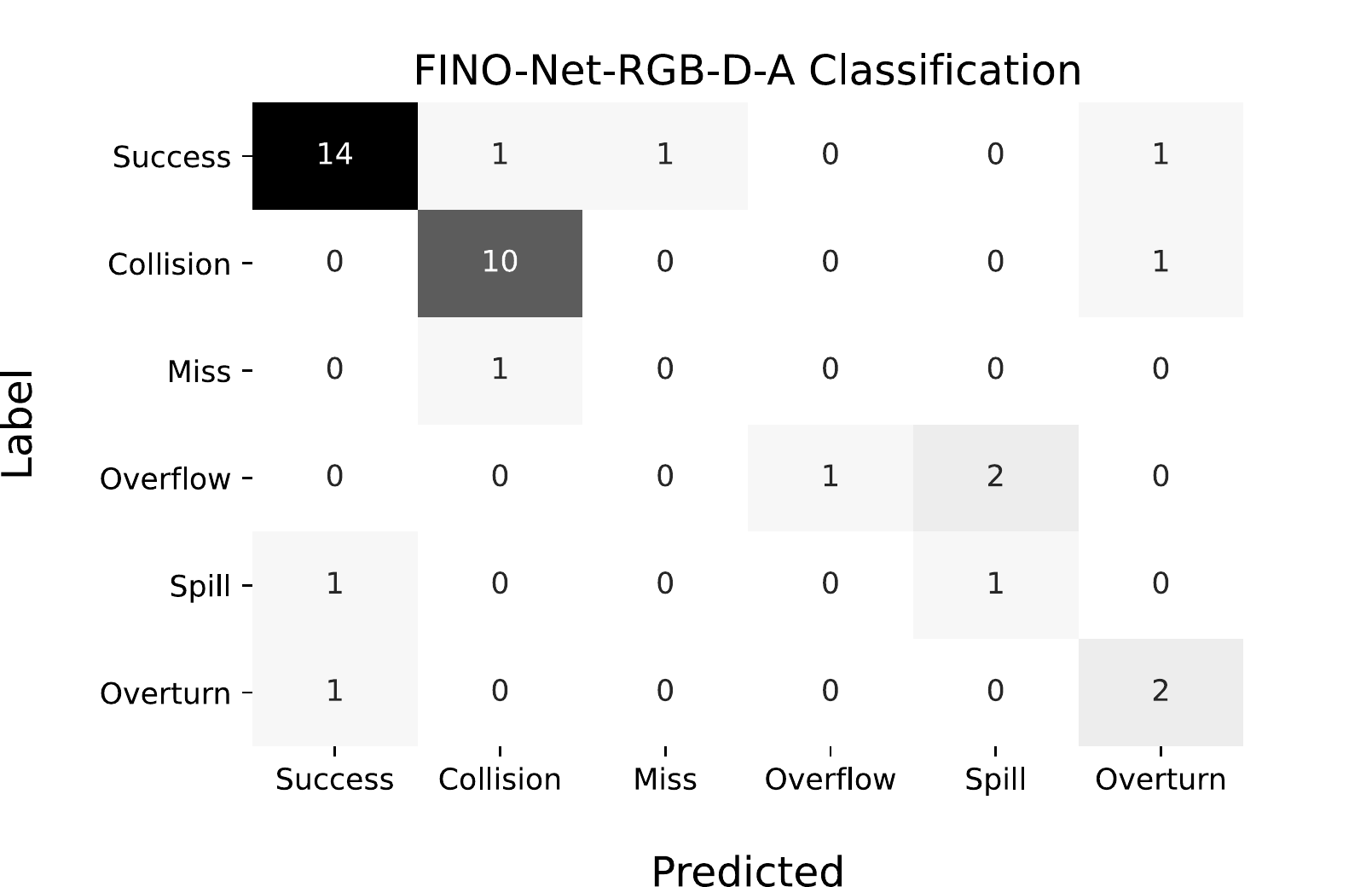}
    \caption{Confusion Matrix for stand-alone FINO-Net-RGB-D-A failure classification. Success class is included. }
    \label{fig:conf_mat6}
\end{figure}

\begin{figure}[t!]
\centering
    \includegraphics[width=\linewidth]{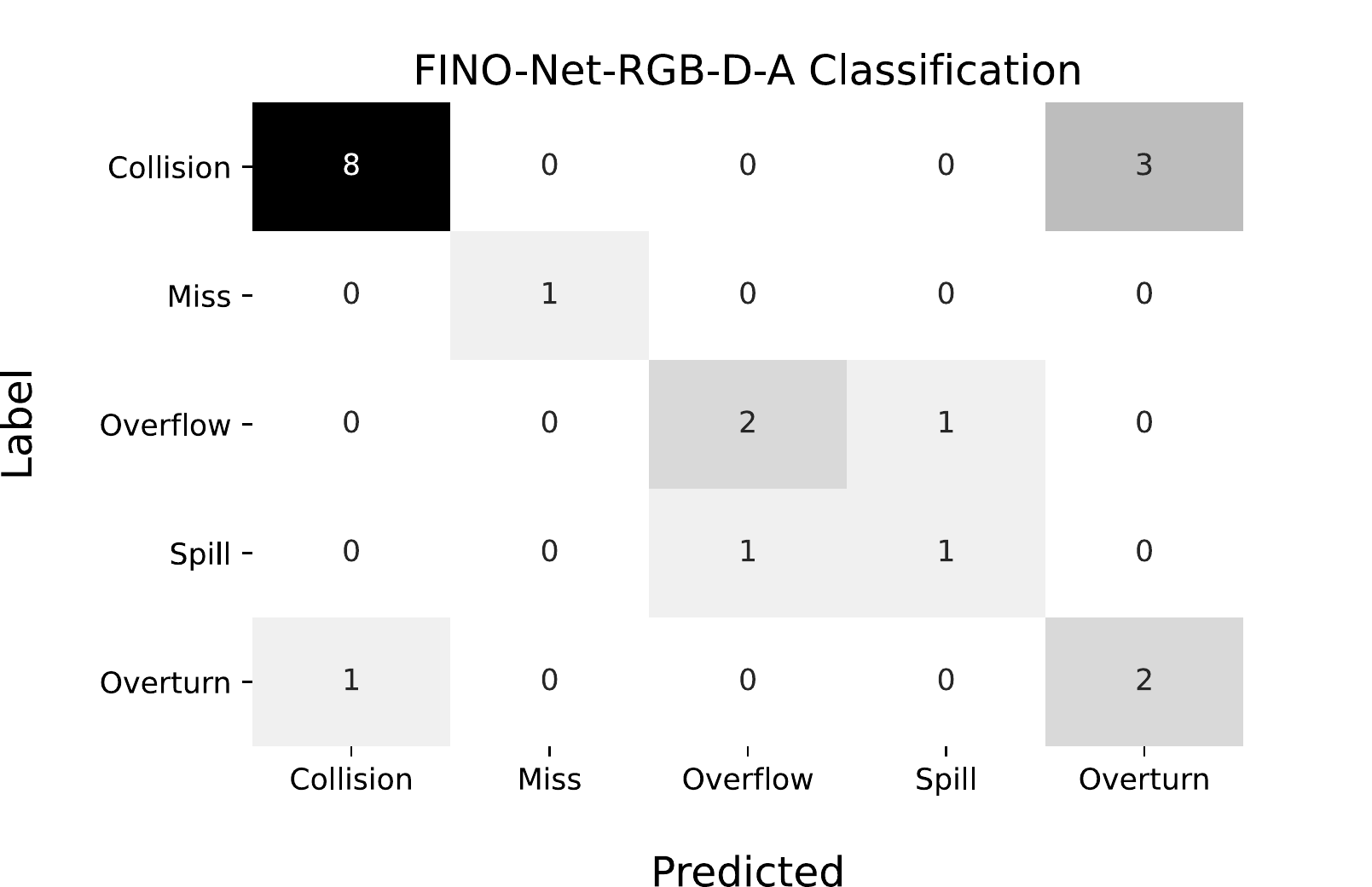}
    \caption{Confusion Matrix for cascaded FINO-Net-RGB-D-A failure classification. Success class is omitted. }
    \label{fig:conf_mat5}
\end{figure}

\subsection{Quantitative Evaluation}
For quantitative assessment, the extended FAILURE dataset is split into training (70\%), validation (10\%) and test sets (20\%). All network weights are initialized randomly and trained for 250 epochs with a learning rate of $1e-5$ using the Adam optimizer. An early stopping strategy is adopted to select the best model based on validation set scores. The test scores obtained with the selected best models are reported in the following subsections.

To prevent overfitting, we augment the data by applying color augmentation and random flipping. For instance, the brightness, contrast, saturation, and hue values of all images in a sequence are randomly changed with a probability of $0.2$. In a similar fashion, each image sequence is flipped vertically with a probability of $0.5$.

FINO-Net is trained with the following inputs for both detection and classification tasks:

\begin{itemize}
    \item FINO-Net-RGB: The network is trained with RGB frames.
    
    \item FINO-Net-D: The network is only trained with the depth (D) frames.
    
    \item FINO-Net-A: Only the audio (A) branch is trained. After the convolutional layers, a single fully connected layer with 64 neurons is applied.
    
    \item FINO-Net-RGB-D: The visual branch of FINO-Net is trained by stacking the RGB and depth frames as input to the network.
    
    \item FINO-Net-RGB-D-A: The entire network is trained with all the given modalities.
\end{itemize}

\begin{figure}[t!]
\centering
    \includegraphics[width=0.8\linewidth]{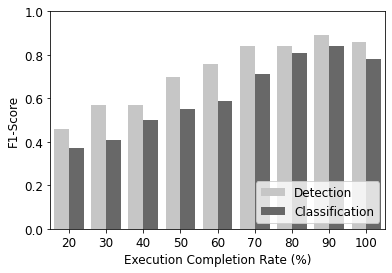}
    \caption{The analysis on real-time demands for failure detection and classification. Different rates of the observation sequence of the execution are provided to FINO-Net-RGB-D-A model}
    \label{fig:data_amount}
\end{figure}

\begin{table}[]
\caption{Effect of frame sampling. Averages of 50 predictions.}
\label{tbl:online_sampling}
\resizebox{\columnwidth}{!}{%
\begin{tabular}{|l|l|c|c|c|}
\hline
          \textbf{Task}                      & \textbf{Network}   & \textbf{Pr} & \textbf{Re} & \textbf{F1} \\ \hline
\multirow{2}{*}{\textbf{Detection}}      & \textbf{FINO-Net-RGBD}  & $0.7331\pm 0.045$         & $0.7276 \pm 0.041$         & $0.7274 \pm 0.041$         \\ \cline{2-5} 
                               & \textbf{FINO-Net-RGBDA} & $0.8201\pm 0.044$         & $0.8092\pm 0.043$         & $\textbf{0.8054}\pm \textbf{0.045}$       \\ \hline
\multirow{2}{*}{\textbf{Classification}} & \textbf{FINO-Net-RGBD}  & $0.6987\pm 0.045$         & $0.6805 \pm 0.049$         & $0.6820 \pm 0.046 $        \\ \cline{2-5} 
                               & \textbf{FINO-Net-RGBDA} & $0.8154\pm 0.039$         & $0.7795 \pm 0.047$         & $\textbf{0.7884} \pm \textbf{0.043}$        \\ \hline
\end{tabular}%
}
\end{table}

\begin{figure*}[t!]
\centering
    \includegraphics[width=\linewidth]{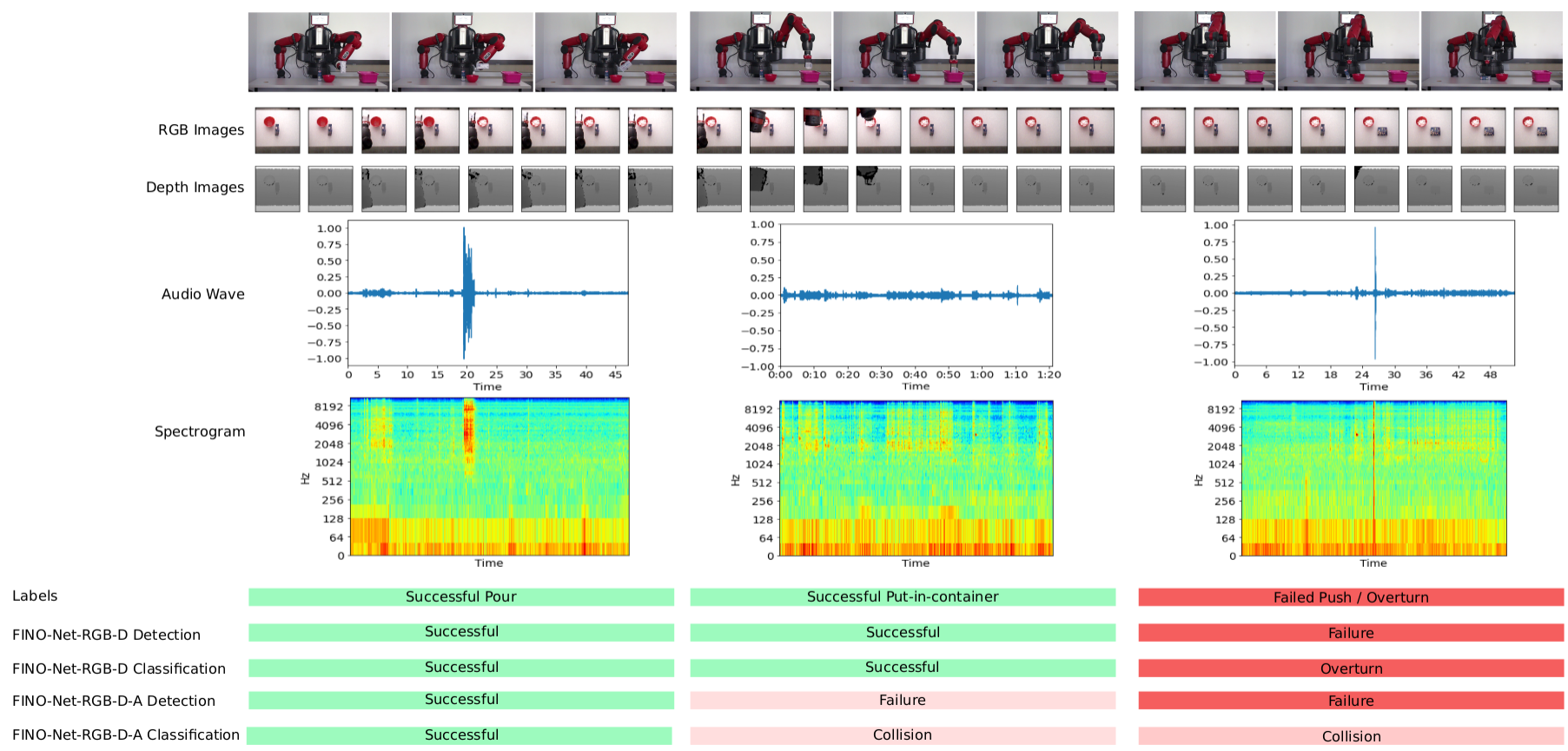}
    \caption{FINO-Net evaluation on a compound scenario. The robot executes a successful pouring, a successful put-in-container, and a failed pushing manipulation action in sequence. From top to bottom, rows present the following: third-person camera view, 8 RGB and depth frame pairs sampled from the execution (i.e., input to FINO-Net), audio wave and corresponding spectrogram images, labels, and predictions. 
    }
    \label{fig:online_detailed}

\end{figure*}

Table \ref{tbl:quant} presents quantitative results. FINO-Net-RGB-D-A is able to achieve 0.8656 for failure detection and 0.7959 F1 scores for failure classification. Results verify that  multimodal sensor fusion  obtains higher scores as visual and auditory modalities provide complementary information.

We have investigated two different approaches for failure type classification. The first, stand-alone approach simultaneously detects and classifies failures. In this approach, the network is trained with both successful and failed executions. There are 6 class labels (i.e., success and failure types). The second, cascaded, approach is also considered, where the failure classification network is only triggered when a failure is detected. For this purpose, we excluded successful executions from the dataset, and only trained the network with failed samples.

Figures \ref{fig:conf_mat6} and \ref{fig:conf_mat5} present confusion matrices for stand-alone and cascaded approaches, respectively. Looking deeper into the results, models confuse classes due to the scene similarities in the post-manipulation phases. For instance, particles are laid on the table after both overflow and spill failures. Primary objects' location and orientation are changed whenever a collision or overturn failure occurs. Even though the causes of the failures are different and different failure indicators are observed in the manipulation phase, similar scene states are observed in the post-manipulation phase.

FINO-Net makes predictions on demand. In order to assess the suitability of FINO-Net for real-time deployment on the robot, we conducted two analyses. First, manipulation is segmented by execution's completion rate, and 8 RGB and depth frame pairs are sampled from the beginning to the end of segment. Figure \ref{fig:data_amount} presents the results of this analysis. Typically, indicators of failure are observed during the second half of the execution. Results verify that the model makes more reliable predictions towards the end of manipulation as more data becomes available to it. Please also note that the models are only trained with sampled frames from the complete manipulation sequence. Yet FINO-Net is capable of detecting and identifying failures in the absence of complete manipulation sequence. 

Model's performance is also affected by the sampled images (i.e., sampled frames may not carry information about the failure phase or the effects of the failure). In order to measure the effect, 50 different image sequences are sampled for each manipulation. Average results are given in Table \ref{tbl:online_sampling}. 

\subsection{Qualitative Evaluation}

We also demonstrate how  FINO-NET behaves on a longer compound scenario (see Figure \ref{fig:online_detailed}). In this scenario, the robot executes a successful pouring, a successful put-in-container, and a failed pushing action in a manipulation sequence order. The scene also contains secondary objects, some of which are novel objects which are not presented in the training phase, as well as task-related objects.

FINO-Net-RGB-D detection and classification models make correct predictions for all three manipulation actions. For the second manipulation action, FINO-Net-RGB-D-A detection and classification models make incorrect predictions as, during the put-in-container phase, 
a sound event is observed (i.e., transparent container collides with purple container). For the final manipulation action, all models except the FINO-Net-RGB-D-A classification model make correct predictions. Even though the model catches the failure, it confuses overturn with collision.

The audio modality is challenging to work with as environmental noises and the robot's ego noise affect the model's performance. Furthermore, most of the object interactions generate a sound event regardless of the manipulation success (e.g., pouring pasta into the pan). Various types of sounds are generated from object-object interactions \cite{Saltali2016}. Even though FAILURE dataset includes a variety of object materials, FINO-Net architecture is able to represent and discriminate successful and failed sound events on these objects.

\section{CONCLUSIONS}

Previous studies investigated model-based and data-driven execution monitoring systems. In our previous works \cite{Inceoglu2018icra, Inceoglu2018iros}, we have also represented auditory and visual inputs as symbolic predicates. However, data-driven manipulation failure detection and classification approaches have been underexplored due to the lack of open-source robot execution datasets. This work bridges the gap by providing FAILURE, 
an open-source, multimodal real-robot execution dataset. In this work, we extend the FAILURE dataset with 99 new multimodal manipulation executions and annotate each with the corresponding failure type. We present, FINO-Net architecture,  an end-to-end frameworkthat directly learns the relationship between inputs and targets. We train the network for both failure detection and failure classification tasks. Quantitative results indicate that FINO-Net is capable of detecting and classifying failures. It also verifies symbolic results, as visual and auditory modalities are complementary to each other and performance is boosted via fusion. As future work, we plan to focus on anticipation of failures before they occur. Thus any potential damage can be minimized and safety can be assured.

\addtolength{\textheight}{-10cm}   




\bibliographystyle{IEEEtran} 
\bibliography{references}

\begin{thebibliography}{10}
\providecommand{\url}[1]{#1}
\csname url@samestyle\endcsname
\providecommand{\newblock}{\relax}
\providecommand{\bibinfo}[2]{#2}
\providecommand{\BIBentrySTDinterwordspacing}{\spaceskip=0pt\relax}
\providecommand{\BIBentryALTinterwordstretchfactor}{4}
\providecommand{\BIBentryALTinterwordspacing}{\spaceskip=\fontdimen2\font plus
\BIBentryALTinterwordstretchfactor\fontdimen3\font minus
  \fontdimen4\font\relax}
\providecommand{\BIBforeignlanguage}[2]{{%
\expandafter\ifx\csname l@#1\endcsname\relax
\typeout{** WARNING: IEEEtran.bst: No hyphenation pattern has been}%
\typeout{** loaded for the language `#1'. Using the pattern for}%
\typeout{** the default language instead.}%
\else
\language=\csname l@#1\endcsname
\fi
#2}}
\providecommand{\BIBdecl}{\relax}
\BIBdecl

\bibitem{Inceoglu2021fino}
A.~Inceoglu, E.~E. Aksoy, A.~C. Ak, and S.~Sariel, ``Fino-net: A deep
  multimodal sensor fusion framework for manipulation failure detection,'' in
  \emph{IEEE/RSJ International Conference on Intelligent Robots and Systems
  (IROS)}, 2021, pp. 6841--6847.

\bibitem{Ersen2017}
M.~{Ersen}, E.~{Oztop}, and S.~{Sariel}, ``Cognition-enabled robot manipulation
  in human environments: Requirements, recent work, and open problems,''
  \emph{IEEE Robotics Automation Magazine}, vol.~24, no.~3, pp. 108--122, 2017.

\bibitem{veruggio2016roboethics}
G.~Veruggio, F.~Operto, and G.~Bekey, ``Roboethics: Social and ethical
  implications,'' \emph{Springer handbook of robotics}, pp. 2135--2160, 2016.

\bibitem{iso_10218:2011_part1}
ISO{~}10218:2011, \emph{Robots and robotic devices -- Safety requirements for
  industrial robots -- Part 1: Robots}.\hskip 1em plus 0.5em minus 0.4em\relax
  ISO, Geneva, Switzerland, 2011.

\bibitem{iso_10218:2011_part2}
ISO{~}10218-2:2011, \emph{Robots and robotic devices -- Safety requirements for
  industrial robots -- Part 2: Robot systems and integration}.\hskip 1em plus
  0.5em minus 0.4em\relax ISO, Geneva, Switzerland, 2011.

\bibitem{morin2011self}
A.~Morin, ``Self-awareness part 1: Definition, measures, effects, functions,
  and antecedents,'' \emph{Social and personality psychology compass}, vol.~5,
  no.~10, pp. 807--823, 2011.

\bibitem{nitsch2013situation}
V.~Nitsch, ``Situation awareness in autonomous service robots,'' 2013.

\bibitem{dos2020situational}
C.~W. Dos~Santos, L.~Nelson~Filho, D.~B. Esp{\'\i}ndola, and S.~S. Botelho,
  ``Situational awareness oriented interfaces on human-robot interaction for
  industrial welding processes,'' \emph{IFAC-PapersOnLine}, vol.~53, no.~2, pp.
  10\,168--10\,173, 2020.

\bibitem{ak2019stop}
A.~C. Ak, A.~Inceoglu, and S.~Sariel, ``When to stop for safe manipulation in
  unstructured environments?'' in \emph{Proceedings of the 18th International
  Conference on Autonomous Agents and MultiAgent Systems}, 2019, pp.
  1767--1769.

\bibitem{Inceoglu2018icra}
A.~Inceoglu, G.~Ince, Y.~Yaslan, and S.~Sariel, ``Comparative assessment of
  sensing modalities on manipulation failure detection,'' in \emph{IEEE ICRA
  Workshop on Perception, Inference and Learning for Joint Semantic, Geometric
  and Physical Understanding}, 2018.

\bibitem{diehl2022did}
M.~Diehl and K.~Ramirez-Amaro, ``Why did i fail? a causal-based method to find
  explanations for robot failures,'' \emph{IEEE Robotics and Automation
  Letters}, vol.~7, no.~4, pp. 8925--8932, 2022.

\bibitem{altan2021went}
D.~Altan and S.~Sariel, ``What went wrong? identification of everyday object
  manipulation anomalies,'' \emph{Intelligent Service Robotics}, vol.~14,
  no.~2, pp. 215--234, 2021.

\bibitem{inceoglu2018violet}
A.~Inceoglu, C.~Koc, B.~O. Kanat, M.~Ersen, and S.~Sariel, ``Continuous visual
  world modeling for autonomous robot manipulation,'' \emph{IEEE Transactions
  on Systems, Man, and Cybernetics: Systems}, vol.~49, no.~1, pp. 192--205,
  2018.

\bibitem{min2019collision}
F.~Min, G.~Wang, and N.~Liu, ``Collision detection and identification on robot
  manipulators based on vibration analysis,'' \emph{Sensors}, vol.~19, no.~5,
  p. 1080, 2019.

\bibitem{xiaoran2021acoustic}
F.~Xiaoran, Y.~Chen, D.~W. Lee, C.~Prepscius, I.~V. Isler, L.~D. Jackel, H.~S.
  Seung, and D.~D. Lee, ``Acoustic collision detection and localization for
  robotic devices,'' Sep.~23 2021, uS Patent App. 17/084,257.

\bibitem{valle2022real}
C.~M.~C. Valle, A.~Kurdas, E.~P. Fortuni{\'c}, S.~Abdolshah, and S.~Haddadin,
  ``Real-time imu-based learning: a classification of contact materials,'' in
  \emph{2022 IEEE/RSJ International Conference on Intelligent Robots and
  Systems (IROS)}.\hskip 1em plus 0.5em minus 0.4em\relax IEEE, 2022, pp.
  1965--1971.

\bibitem{Saltali2016}
I.~Saltali, S.~Sariel, and G.~Ince, ``Scene analysis through auditory event
  monitoring,'' in \emph{Proceedings of the International Workshop on Social
  Learning and Multimodal Interaction for Designing Artificial Agents}.\hskip
  1em plus 0.5em minus 0.4em\relax ACM, 2016, p.~5.

\bibitem{haddadin2017robot}
S.~Haddadin, A.~De~Luca, and A.~Albu-Sch{\"a}ffer, ``Robot collisions: A survey
  on detection, isolation, and identification,'' \emph{IEEE Transactions on
  Robotics}, vol.~33, no.~6, pp. 1292--1312, 2017.

\bibitem{proper2023aim}
B.~Proper, A.~Kurdas, S.~Abdolshah, S.~Haddadin, and A.~Saccon, ``Aim-aware
  collision monitoring: Discriminating between expected and unexpected
  post-impact behaviors,'' 2023.

\bibitem{Mandlekar2018roboturk}
A.~Mandlekar, Y.~Zhu, A.~Garg, J.~Booher, M.~Spero, A.~Tung, J.~Gao, J.~Emmons,
  A.~Gupta, E.~Orbay \emph{et~al.}, ``Roboturk: A crowdsourcing platform for
  robotic skill learning through imitation,'' in \emph{Conference on Robot
  Learning}, 2018, pp. 879--893.

\bibitem{Mandlekar2019roboturk}
A.~Mandlekar, J.~Booher, M.~Spero, A.~Tung, A.~Gupta, Y.~Zhu, A.~Garg,
  S.~Savarese, and L.~Fei-Fei, ``Scaling robot supervision to hundreds of hours
  with roboturk: Robotic manipulation dataset through human reasoning and
  dexterity,'' in \emph{IEEE/RSJ International Conference on Intelligent Robots
  and Systems (IROS)}, 2019, pp. 1048--1055.

\bibitem{Dasari2019robonet}
S.~Dasari, F.~Ebert, S.~Tian, S.~Nair, B.~Bucher, K.~Schmeckpeper, S.~Singh,
  S.~Levine, and C.~Finn, ``Robonet: Large-scale multi-robot learning,'' in
  \emph{CoRL 2019: Volume 100 Proceedings of Machine Learning Research}, 2019.

\bibitem{Levine2018learning}
S.~Levine, P.~Pastor, A.~Krizhevsky, J.~Ibarz, and D.~Quillen, ``Learning
  hand-eye coordination for robotic grasping with deep learning and large-scale
  data collection,'' \emph{The International Journal of Robotics Research},
  vol.~37, no. 4-5, pp. 421--436, 2018.

\bibitem{Finn2016unsupervised}
C.~Finn, I.~Goodfellow, and S.~Levine, ``Unsupervised learning for physical
  interaction through video prediction,'' in \emph{Advances in Neural
  Iinformation Processing Systems (NeurIPS)}, 2016, pp. 64--72.

\bibitem{Chalapathy2019}
R.~Chalapathy and S.~Chawla, ``Deep learning for anomaly detection: A survey,''
  \emph{arXiv preprint arXiv:1901.03407}, 2019.

\bibitem{Gertler1998}
J.~Gertler, \emph{Fault detection and diagnosis in engineering systems}.\hskip
  1em plus 0.5em minus 0.4em\relax CRC press, 1998.

\bibitem{Fritz2005}
C.~Fritz, ``Execution monitoring -- a survey,'' University of Toronto, Tech.
  Rep., 2005.

\bibitem{Pettersson2005}
O.~Pettersson, ``Execution monitoring in robotics: A survey,'' \emph{Robotics
  and Autonomous Systems}, vol.~53, no.~2, pp. 73--88, 2005.

\bibitem{Pettersson2007}
O.~Pettersson, L.~Karlsson, and A.~Saffiotti, ``Model-free execution monitoring
  in behavior-based robotics,'' \emph{Trans. on Systems, Man, and Cybernetics,
  Part B: Cybernetics}, vol.~37, no.~4, pp. 890--901, 2007.

\bibitem{Mauro2019}
L.~Mauro, F.~Puja, S.~Grazioso, V.~Ntouskos, M.~Sanzari, E.~Alati, and
  F.~Pirri, ``Visual search and recognition for robot task execution and
  monitoring,'' \emph{arXiv preprint arXiv:1902.02870}, 2019.

\bibitem{Sathish2019}
V.~Sathish, M.~Orkisz, M.~Norrlof, and S.~Butail, ``Data-driven gearbox failure
  detection in industrial robots,'' \emph{IEEE Transactions on Industrial
  Informatics}, 2019.

\bibitem{Zhou2020}
X.~Zhou, H.~Wu, J.~Rojas, Z.~Xu, and S.~Li, \emph{Nonparametric Bayesian Method
  for Robot Anomaly Monitoring}.\hskip 1em plus 0.5em minus 0.4em\relax
  Springer Singapore, 2020, pp. 51--93.

\bibitem{Wu2019}
H.~Wu, Y.~Guan, and J.~Rojas, ``A latent state-based multimodal execution
  monitor with anomaly detection and classification for robot introspection,''
  \emph{Applied Sciences}, vol.~9, no.~6, p. 1072, 2019.

\bibitem{Park2018}
D.~Park, Y.~Hoshi, and C.~C. Kemp, ``A multimodal anomaly detector for
  robot-assisted feeding using an lstm-based variational autoencoder,''
  \emph{Robotics and Automation Letters}, vol.~3, no.~3, pp. 1544--1551, 2018.

\bibitem{Park2019}
D.~Park, H.~Kim, and C.~C. Kemp, ``Multimodal anomaly detection for assistive
  robots,'' \emph{Autonomous Robots}, vol.~43, no.~3, pp. 611--629, 2019.

\bibitem{thoduka2021}
S.~Thoduka, J.~Gall, and P.~G. Pl{\"o}ger, ``Using visual anomaly detection for
  task execution monitoring,'' in \emph{International Conference on Intelligent
  Robots and Systems (IROS)}, 2021, pp. 4604--4610.

\bibitem{gohil2022}
P.~Gohil, S.~Thoduka, and P.~G. Pl{\"o}ger, ``Sensor fusion and multimodal
  learning for robotic grasp verification using neural networks,'' in
  \emph{2022 26th International Conference on Pattern Recognition
  (ICPR)}.\hskip 1em plus 0.5em minus 0.4em\relax IEEE, 2022, pp. 5111--5117.

\bibitem{altan2022}
D.~Altan and S.~Sariel, ``Clue-ai: A convolutional three-stream anomaly
  identification framework for robot manipulation,'' \emph{Preprint
  arXiv:2203.08746}, 2022.

\bibitem{Inceoglu2018iros}
A.~Inceoglu, G.~Ince, Y.~Yaslan, and S.~Sariel, ``Failure detection using
  proprioceptive, auditory and visual modalities,'' in \emph{IEEE/RSJ
  International Conference on Intelligent Robots and Systems (IROS)}.\hskip 1em
  plus 0.5em minus 0.4em\relax IEEE, 2018, pp. 2491--2496.

\bibitem{Endsley1995situation}
M.~R. Endsley, ``Toward a theory of situation awareness in dynamic systems,''
  \emph{Human Factors}, vol.~37, no.~1, pp. 32--64, 1995.

\end{thebibliography}

\end{document}